\documentclass[a4paper]{./styles/svproc}
 \usepackage{algpseudocode}

\usepackage[T1]{fontenc}
\usepackage{graphicx}
\usepackage{amsmath, amssymb}
\usepackage{booktabs}
\usepackage{hyperref}
\usepackage{xcolor}
\usepackage{subcaption}
\usepackage{multirow}
\usepackage{graphicx}   
\usepackage{url}

\begin{document}
\mainmatter              

\newcommand{\todo}[1]{\textcolor{red}{[TODO: #1]}}

 \newcommand{\Cframe}{\mathcal{C}}      
 \newcommand{\Aframe}{\mathcal{A}}      
\newcommand{\Bframe}{\mathcal{B}}      
\newcommand{\Mframe}{\mathcal{M}}      
\newcommand{\SO}[1]{\mathrm{SO}(#1)}
\newcommand{\SE}[1]{\mathrm{SE}(#1)}
 
\title{JOIN: Anchor-Grasp-Conditioned \\Joining via Opposition, Inference, and Navigation for Bimanual Assistive Manipulation}

\titlerunning{Joining via Opposition, Inference, and Navigation}  
%

\author{Drake Moore \and Matt Cheng\and Xiang Zhi Tan$^{\dagger}$ \and Ta\c{s}k{\i}n Pad{\i}r$^{\dagger\star}$}%
\authorrunning{Moore et al.}
\tocauthor{Drake Moore}
\institute{Northeastern University, Boston MA 02115, USA\\
\email{moore.dr@northeastern.edu}}

\maketitle

\renewcommand{\thefootnote}{\fnsymbol{footnote}}
\footnotetext[4]{Denotes equal advising.}
\footnotetext[1]{Ta\c{s}k{\i}n Pad{\i}r holds concurrent appointments as a Professor of Electrical and Computer Engineering at Northeastern University and as an Amazon Scholar. This paper describes work performed at Northeastern University and is not associated with Amazon.}
\begin{abstract}

Assistive mobility and manipulation platforms have received increasing attention as a means of restoring independence to individuals with disabilities. While effective for many basic activities of daily living (ADLs), a significant percentage of everyday tasks such as opening a jar, pouring a liquid, lifting a tray, or basic meal preparation, is fundamentally bimanual and remains out of reach for any single-arm system. Adding a second arm to a wheelchair is impractical, due to the additional power draw, cost, and the loss of space required for transfers and mobility. We instead propose a heterogeneous, on-demand bimanual system, in which a wheelchair-mounted \emph{anchor} arm is joined when needed by a summoned mobile manipulator that serves as a \emph{complement} arm. The central technical problem, which we call \emph{bimanual joining}, is conditional: the anchor has already committed to a grasp, and the complement arm must choose where to stand and what to grasp to complete the task. We formulate bimanual joining as a three-phase decomposition (plan, drive, grasp) and show that a vision-language model (VLM), coupled with standard geometric tools, provides task-level knowledge sufficient to solve a representative class of bimanual ADLs. Our system \textbf{JOIN}, contributes (i) a wheelchair-referenced opposition score, and (ii) task-conditioned directional manipulability.
We evaluate JOIN on a Kinova Gen3 anchor and a Hello Robot Stretch~3 complement on representative same-object and different-object tasks. JOIN accomplished more attempts (19/20) than state-of-the-art methods (14/20) and required markedly less correction by the operator.

\end{abstract}
 

\section{Introduction}
\label{sec:intro}
 
Wheelchair-mounted manipulators have been shown to significantly enhance the independence and autonomy of individuals\cite{jenamani2025enhancing,padir2015personalized}.
However, the single arm setup is inherently limiting -- it cannot perform the large class of activities of daily living (ADLs) that require two hands acting in coordination.
For example, opening a water bottle requires one hand to stabilize the body and another to twist the lid. Other tasks, such as lifting a large tray, require both arms to be at opposite ends, coordinating their movements.
Pouring a drink, cutting a fruit, and opening a bottle are all routine tasks that a wheelchair user may encounter, but that a standard robotic wheelchair with a single arm cannot complete easily.

The most straightforward solution is to mount a second robotic arm directly on the wheelchair~\cite{wang2012permma,gandhi2025bimanualshared}.
While this greatly expands the robotic wheelchair's capability, it is mechanically and ergonomically costly -- adding weight, increasing power draw, and occupying space that the wheelchair users need for transfers and mobility.
It also keeps an entire manipulator idle whenever the user's task happens to be unimanual -- a latent cost that an on-demand complement avoids by performing other tasks elsewhere in the environment when not summoned.
\begin{figure}[t]
    \centering
    \includegraphics[width=\textwidth]{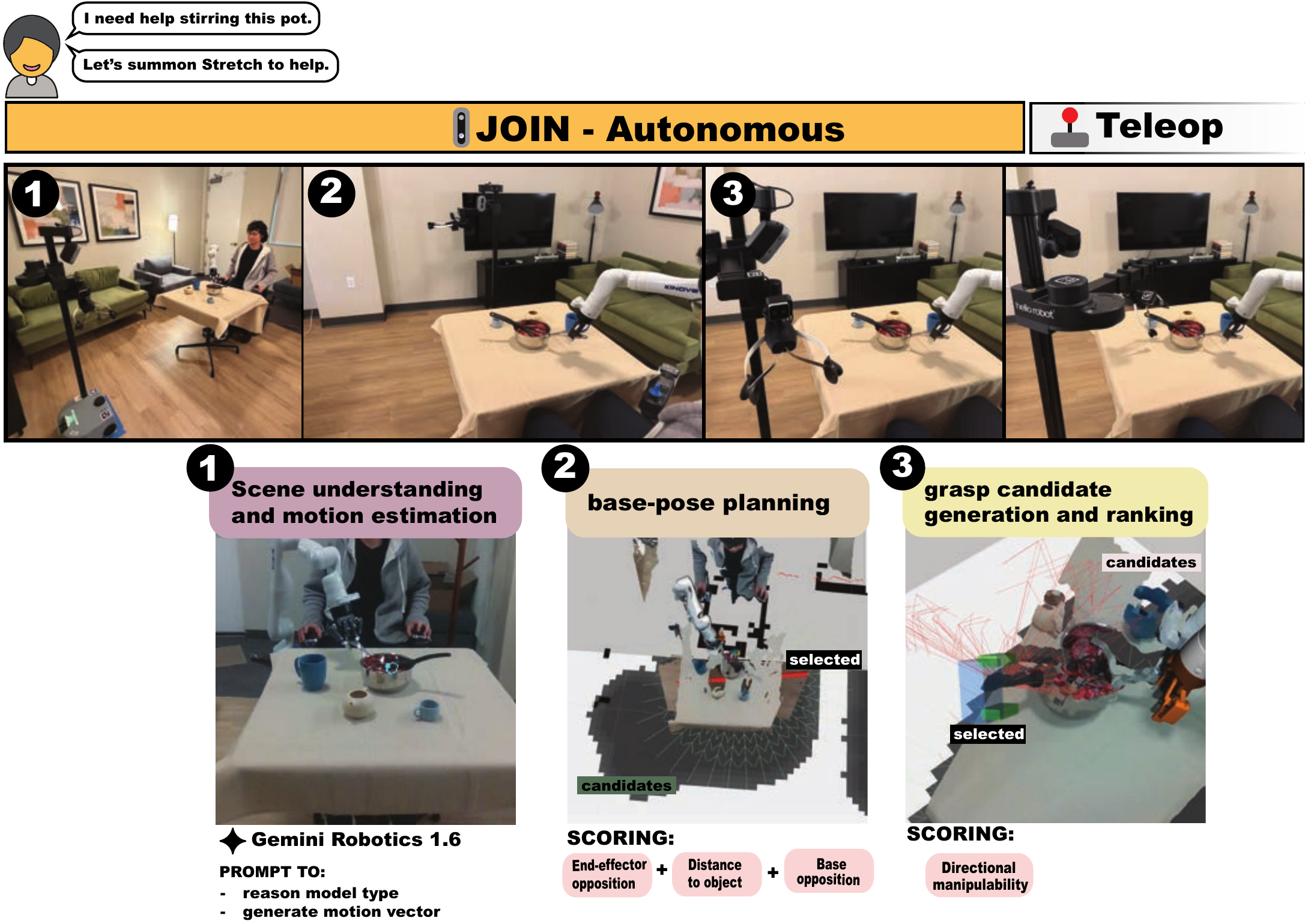}
    \caption{Overview of the full JOIN system in action. \textbf{1:} The complement first reasons over the required task, the anchor's current grasp, and the trajectory the target object will take to produce a target and motion vectors. \textbf{2:} the complement selects the joining location, scored by opposition and proximity, and navigates to it. \textbf{3:} The complement samples grasps for the target object, and selects the one with the highest task-based manipulability.}
    \label{fig:teaser}
\end{figure}

Instead of augmenting the robotic wheelchair further, we explored how it can leverage other robots in the environment, such as a mobile manipulator, to join in and allow the user to operate its arm as the wheelchair's second arm.
In this setup, a wheelchair-mounted arm (the \emph{anchor}) initiates a task and commits to a grasp. 
A mobile manipulator (the \emph{complement}) is summoned to the scene and contributes a second hand.
When the task completes, it can leave the environment.
Instead of having the user teleoperate the robot all the way, we investigate how the complement robot can reason about the task and pick an optimal joining location.
We call this problem -- \emph{bimanual joining}: given an anchor arm mid-task, determine where the complement arm should position itself and what grasp it should execute, so that they together can complete the task.

Bimanual joining is structurally different from traditional fixed bimanual robot control along two axes. 
First, it is conditional: the anchor has already committed to a grasp, so the anchor's location, orientation, and approach direction are inputs to the complement's decision rather than co-decided with it.
Second, the complement is on a mobile base away from the task and with a robot arm that has a potentially different kinematic structure from the anchor.
This introduces a base-placement subproblem not present in the fixed bimanual setting: the complement must select a position that affords easy task completion while aligning with the wheelchair arm, easing the operator's task.
First, the complement must reason where to grasp: given a jar held by its body, the complementary grasp is on the lid; given a kettle about to pour, the complementary grasp is the cup, oriented for receiving.
The second is which grasp is good for the upcoming motion the task requires: unscrewing a lid demands rotational capacity about the jar's axis; lifting a tray demands vertical linear capacity.



To solve the Bimanual joining problem, we propose a novel system -- Joining via Opposition, Inference, and Navigation (JOIN).
JOIN is structured as three phases: \emph{(1)plan}, in which the complement robot captures a workspace view and utilizes a vision-language model (VLM) to infer the \emph{what} should be grasped and \emph{how} the grasped object will move under the task. 
\emph{(2) drive}, in which candidate base poses are sampled and scored by how they complement the anchor arm.
\emph{(3) \emph{grasp}}, in which grasp candidates from the close-range view are selected by how easily the grasp can express the task-direction inferred by the VLM.  
\\
\noindent
This paper has the following contributions:
\begin{enumerate}
     \item We introduce a new formalism to the \textbf{bimanual joining problem}: a conditional form of bimanual manipulation in which one arm has committed to a grasp and the other must complement it, arriving from an independent mobile base. 
    \item We propose a \textbf{three-phase decomposition} --- plan, drive, grasp --- in which the anchor grasp conditions all three phases and the VLM is invoked only for task-level reasoning, with all 3-D quantities computed geometrically.
    \item We introduce a \textbf{anchor-referenced opposition score} for complement base-pose selection that prefers placements consistent with the natural geometry of human bimanual collaboration, at both the gripper and the body.
    \item We introduce \textbf{task-conditioned directional manipulability}: the VLM annotates the post-grasp motion the task requires, and candidate grasps are scored by the complement's manipulability along that motion rather than by an isotropic measure, preferring grasps that are dexterous for the motion the task actually requires.
    \item We evaluated \textbf{JOIN} on real hardware --- a Kinova Gen3 wheelchair-mounted anchor and a Hello Robot Stretch~3 complement across four bimanual tasks. JOIN is also embodiment agnostic. The system can be extended beyond the evaluated hardware, including anchors not mounted on a wheelchair. 

\end{enumerate}
JOIN contributes two methodological components that we believe generalize to other heterogeneous bimanual applications: an anchor-referenced opposition score that places the complement across the target from the anchor arm, and task-conditioned directional manipulability, in which the task direction is obtained automatically from a robotics VLM's trajectory annotation

\section{Related Work}
\label{sec:related}
 
\subsection{Bimanual manipulation.}
Bimanual manipulation has been studied extensively on fixed-base homogeneous dual-arm platforms, where numerous benchmarks and policies have demonstrated the capabilities of bimanual coordination~\cite{grotz2024peract2,jiang2025decoupledbimanual,im2025twinvla}. Some related work has examined bimanual heterogeneous platforms using LLM- and VLM-driven multi-robot frameworks to coordinate agents of differing embodiments~\cite{chen2025emos,mandi2024roco}, and some prior work has one arm reconfigure the scene to enable the other's manipulation~\cite{shen2026bipremanip}. These systems share a calibrated workspace and a planner that decides both arms together. Bimanual joining differs on each count: the two arms arrive from independent mobile bases with different kinematics, and one arm's grasp is fixed before the other's decision begins.
 
\subsection{Mobile manipulation and base placement.}
Mobile manipulation of everyday objects in domestic environments has been addressed with full system architectures integrating navigation, perception, and grasping~\cite{kelecstemur2019system,liu2024okrobot}. Where to position the base is classically handled by precomputing the arm's reachable workspace as a reachability or capability map and inverting it to recover base poses from which a target is reachable~\cite{zacharias2007capability,vahrenkamp2013reachabilityinversion}, or as a learned behavior prior~\cite{jauhri2022reachabilitypriors}. Base placement for a single mobile manipulator has thus been studied extensively, but the problem shifts when the base pose must also be consistent with a second, independently placed arm: reachability becomes a hard constraint we combine with a wheelchair-referenced opposition preference and a task-directional manipulability criterion obtained automatically from a VLM rather than the sole objective.

\subsection{Assistive manipulation for wheelchair users.}
Physically assistive robots have shown to be able to restore independence in activities of daily living (ADLs)~\cite{nanavati2024assistivesurvey}, with systems spanning robot-assisted feeding~\cite{jenamani2024flairfeeding} and wearable-interface teleoperation of mobile manipulators~\cite{padmanabha2024wearableteleop}. However, uch systems almost universally assume a single arm acting on the user's behalf, which limits the user's ability to perform bimanual ADLs. Some prior works have mounted a second arm directly on the chair~\cite{wang2012permma,gandhi2025bimanualshared}; but this incurs the weight, power, and footprint costs, motivating our on-demand formulation. A related line of work coordinates multiple assistive devices as cooperating agents for a single ADL~\cite{ye2025cartmpc}. We share the multi-agent, physically coupled framing but address open-space bimanual joining rather than a single instrumented transfer task. To our knowledge no prior wheelchair system summons an independent mobile manipulator on demand to supply the second hand.

\subsection{VLMs for manipulation.}
A growing body of work uses vision-language models to inject task semantics into manipulation, both by predicting bimanual affordance regions and jointly allocating arms on fixed dual-arm platforms~\cite{hahne2026taskaware,heidinger2025handedafforder,chen2026vlmsfd} and by grounding open-vocabulary objects for mobile pick-and-place~\cite{liu2024okrobot,liu2025dynamem,shi2025hirobot}. A lesson we adopt is that VLMs are reliable for \emph{selecting} among, or grounding to, geometrically generated candidates\cite{liu2024okrobot}. Concurrent work by Hahne et al.~\cite{hahne2026taskaware} frames bimanual manipulation as a joint affordance-localization and arm-allocation problem, and uses a VLM to filter 6-DoF grasp candidates on a fixed dual-arm platform. Our work differs along three axes that collectively define a different problem:

 
 
\section{Problem Formulation}
\label{sec:problem}
In this section, we formalize \emph{bimanual joining} and identify the structural properties of the problem. 

\subsection{Bimanual Joining}
\label{sec:problem:definition}

Let $A$ denote the anchor arm, rigidly mounted on the user's wheelchair, and $C$ the complement mobile manipulator. At some time $t_0$, the anchor has committed to a grasp $g_A = (o_A, \mathbf{T}_A)$, where $o_A$ is the grasped object (e.g., mug handle) as recognized by the VLM, and $\mathbf{T}_A \in \SE{3}$ is the gripper pose in the world frame. The anchor is associated with a natural-language task description $\tau$ (e.g., \emph{``pour water from the bottle into the cup''}).

The goal of the \emph{bimanual joining problem} is to find a complementary grasp $g_C = (o_C, p_C, \mathbf{T}_C)$ where, together with $g_A$, makes $\tau$ achievable.
Since the complementary robot may be far away, we would also need to find a complement base pose $\mathbf{q} \in \SE{2}$ from which $\mathbf{T}_C$ is reachable, collision-free with respect to $A$ and the environment, and compatible with the complement arm's sensor field of view during execution.


We explicitly define the target object for each grasp, as the bimanual tasks may involve the same or different objects.
For example, tasks such as opening a jar (anchor holds body, complement grasps lid), opening a bottle, or holding a tray stably, are in the \emph{same-object mode}, requiring both arms to grasp the same physical object ($o_C = o_A$).
The alternate, \emph{different-object mode}, has the two arms grasping different task-related objects ($o_C \neq o_A$), such as in pouring (anchor holds carton, complement holds cup) or serving (anchor holds ladle, complement holds bowl).
The mode, target object, and grasped partare inferred from $\tau$ by a VLM. The inferred grasped part is then converted into a 3-D target $\mathbf{p}^*$ using the depth image. 
The VLM additionally emits a task-motion characterization---the dominant linear direction and/or rotation axis the manipulated object will undergo, which complement manipulability is later evaluated (Section~\ref{sec:method}). 


To achieve this, we decompose the joining task into three phases. First, the robot must approach the environment to a location that has a full view of the workspace in order to reason over the existing grasp, and any task-relevant objects or environmental features. Next, given this understanding of the task, it must drive to the location that best enables it to perform the collaborative grasp and actions. Lastly, after it is in the best base location, perform the optimal grasp given the constraints/requirements of the task.

\subsection{The Anchor Grasp as a Key Constraint}
\label{sec:problem:anchor}

The anchor grasp constrains the complement arm's subsequent decisions. The complement's actions $\mathbf{q}$ must respect (i) a wheelchair-referenced opposition preference (Section~\ref{sec:method}), and (ii) a task-directional manipulability criterion, evaluated at $\mathbf{T}_C$ against the VLM-supplied task motion, quantifying how well the complement can move along the task direction (Section~\ref{sec:method}). 

\section{Method}
\label{sec:method}
 \begin{figure}[t]
    \centering
    \includegraphics[width=\textwidth]{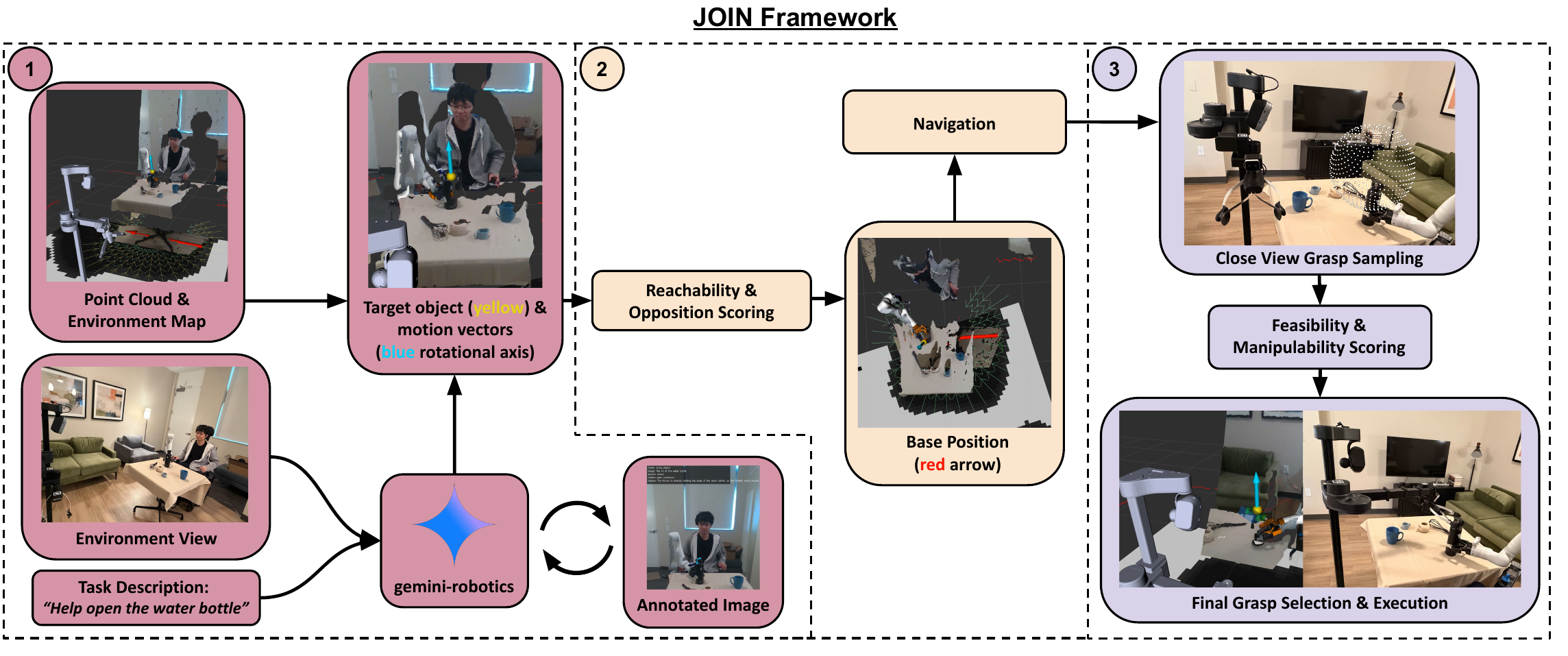}
    \caption{The three-phase JOIN framework. \textbf{(1)~Plan:} a VLM identifies the target object and motion vectors from the environment view and task description. \textbf{(2)~Drive:} base poses are scored by reachability and opposition, and the complement navigates to the selected pose. \textbf{(3)~Grasp:} viable candidates from a close-range view are ranked directional manipulability, and the best is executed.}    
    \label{fig:system}
\end{figure}

When bimanual capability is needed by the user, the user summons the complement robot to a predefined location in the environment that enables it to have a full view of the environment. 
We envision these locations to be at the edge of different living spaces and leave it to future work to further optimize the positioning.
Once the robot arrived at the predefined location, the user then triggers JOIN with a task description, $\tau$.
 
JOIN is structured as three phases corresponding to two viewpoints. Phase~1 (\emph{plan}) uses a fixed wide-angle observation pose to interpret the task and produce a 6-DoF motion prior. Phase~2 (\emph{drive}) plans a base pose $\mathbf{q} \in \SE{2}$ that places the complement arm in a location naturally suited for a collaborative grasp. Phase~3 (\emph{grasp}) generates and ranks grasp candidates from that close-range view. As the system needs to reason from different perspectives, we use the following frame conventions in the paper. $\Cframe$ denotes the complement-arm head-camera frame; $\Bframe$ denotes the complement base frame; $\Aframe$ denotes the anchor base frame.

\subsection{Phase 1: scene understanding and motion estimation}
\label{sec:phase1}
\begin{figure}[t]
    \centering
    \includegraphics[width=\textwidth]{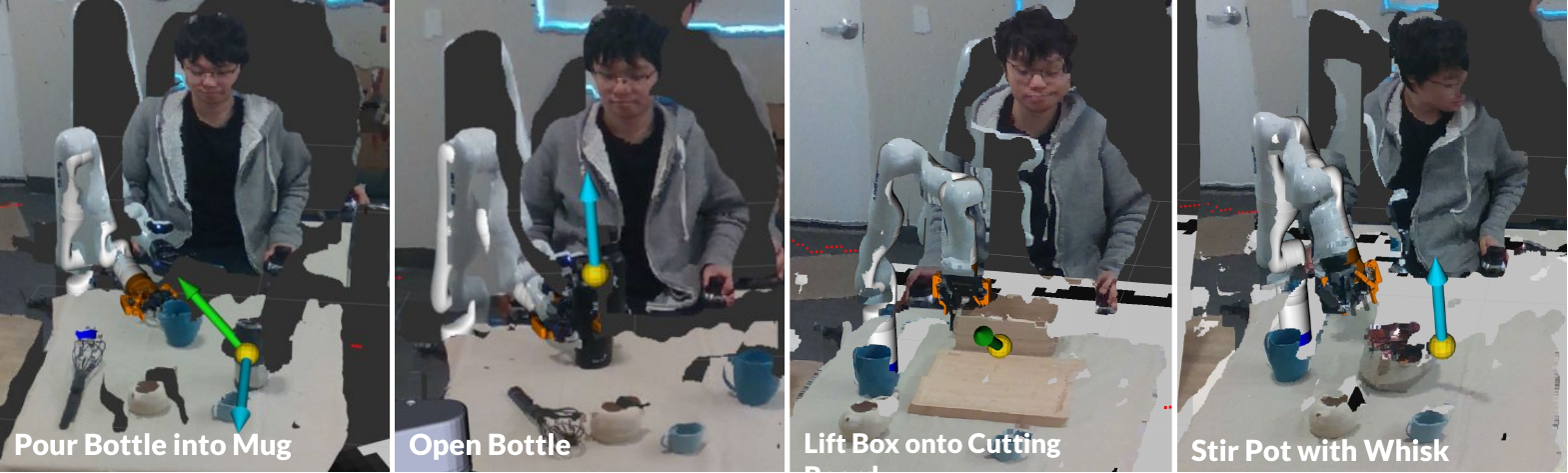}
    \caption{Examples of the VLM-generated motions for each of the tasks. Green arrows indicate the linear direction of the task, and blue arrows indicate the rotational axes.}
    \label{fig:twist}
\end{figure}

Phase~1 enables the robot to reason about the task, required objects, and motion required to completed the task described in $\tau$. 
A head-camera image $I$ and task description $\tau$ are given to a prompted spatially-grounded robotics VLM (Gemini Robotics-ER 1.6 model~\cite{gemini-robotics-er}) to reason the following features in a single call:
\begin{itemize}
  \item a discrete mode $m \in \{\textsc{same\_object}, \textsc{different\_object}\}$.
  \item a target pixel $\mathbf{u}^* \in I$ marking the complement grasp contact region, and a richer textual description $\ell^+$ of that target retained for the Phase~3 selection prompt (e.g.\ \emph{``the opposite rim of the wooden box''}).
  \item a \emph{motion type} $\mu \in \{\textsc{linear}, \textsc{mixed\_linear}, \textsc{balanced}, \textsc{mixed\_rotational}, \\\textsc{rotational}\}$ characterizing the main kinematic motions of the task. 
  \item a unit linear vector ${}^{\Cframe}\hat{\mathbf{d}}_v \in S^2$, interpreted as the primary translational direction of the manipulated object if $\mu \neq \textsc{rotational}$.
  \item a unit angular-axis vector ${}^{\Cframe}\hat{\mathbf{d}}_\omega \in S^2 \cup \{\varnothing\}$ describing the primary rotation axis if $\mu \neq \textsc{linear}$.
\end{itemize}
The motion vectors are produced \emph{zero-shot from the model's spatial reasoning over $I$}, a capability demonstrated for the Gemini Robotics-ER 1.6 model we use~\cite{gemini-robotics-er}.
The generated vectors for our tasks are shown in Fig~\ref{fig:twist}. 
As a chain-of-thought intermediate, the model also produces a polyline tracing the manipulated object's trajectory through $I$; this is used for visualization/reasoning and is not consumed by the geometric pipeline. The central architectural choice of Phase~1 is that the VLM expresses task-level motion semantics directly as camera-frame vectors and a categorical motion type, which the downstream pipeline consumes as priors. The 3-D complement target is recovered directly from depth values centered on the target pixel $\mathbf{u}^*$.

\subsection{Phase 2: base-pose planning}
\label{sec:phase2}

Phase~2 plans a complement base pose $\mathbf{q} = (x, y, \theta) \in \SE{2}$ satisfying two criteria the anchor grasp imposes: a close-range, unoccluded view of the target sufficient for dense grasp synthesis, and opposition geometry referenced to the wheelchair, so the complement platform sits consistently with natural bimanual collaboration. Candidates are sampled on a discretized $\SE{2}$ grid around the target, and only collision-free samples are kept.

\subsubsection{Scoring:}
Each surviving candidate $\mathbf{q}$ is scored by a weighted sum of three normalized terms,
\begin{equation}
  S(\mathbf{q})
    = w_d\, S_d(\mathbf{q})
    + w_{o,\text{ee}}\, S_{o,\text{ee}}(\mathbf{q})
    + w_{o,\text{base}}\, S_{o,\text{base}}(\mathbf{q}),
  \label{eq:phase2_score}
\end{equation}
with $(w_d, w_{o,\text{ee}}, w_{o,\text{base}}) = (0.30, 0.35, 0.35)$. 
We keep the weights static rather than claiming optimality, leaving their tuning to future work.

\emph{Distance} ($S_d$) is a Gaussian on the complement-to-target distance $\delta(\mathbf{q})$, centered at the middle of the Stretch robot's reachable range:
\begin{equation}
  S_d(\mathbf{q}) = \exp\!\left(
    -\frac{(\delta(\mathbf{q}) - \delta^*)^2}{\sigma_d^2}
  \right),
  \quad \delta^* = 0.6~\mathrm{m},\; \sigma_d = 0.20~\mathrm{m}.
\end{equation}


\emph{End-effector opposition} ($S_{o,\text{ee}}$) prefers candidates from which the complement end-effector approaches the target from the opposite side of the anchor end-effector. Working in the workspace $XY$ plane, with $\mathbf{p}_\text{ee}$ the projected complement end-effector position at $\mathbf{q}$ and $\mathbf{p}_\text{ee}^A$ the anchor end-effector projection,

\begin{equation}
  S_{o,\text{ee}}(\mathbf{q}) = \tfrac{1}{2}\!\left(
    1 +
    \frac{(\mathbf{p}^* - \mathbf{p}_\text{ee}^A)\cdot
          (\mathbf{p}_\text{ee} - \mathbf{p}^*)}
         {\|\mathbf{p}^* - \mathbf{p}_\text{ee}^A\|\,
          \|\mathbf{p}_\text{ee} - \mathbf{p}^*\|}
  \right) \in [0, 1].
\end{equation}
 
\emph{Base opposition} ($S_{o,\text{base}}$) prefers complement base placements that are similar to the anchor's base \emph{mirrored across the anchor's forward axis}, reproducing the symmetric stance of a standard bimanual platform. Expressing the target in the anchor base rame $\Aframe$ ($+X$ forward) and reflecting across that axis gives the mirrored anchor-base-to-target vector ${}^{\Aframe}\tilde{\mathbf{v}} = \mathrm{diag}(1, -1, 1)\, {}^{\Aframe}\mathbf{p}^*$, transformed to the workspace plane as $\tilde{\mathbf{v}}$. The score is its cosine  similarity with the actual base-to-target vector:

\begin{equation}
  S_{o,\text{base}}(\mathbf{q}) = \tfrac{1}{2}\!\left(
    1 +
    \frac{\tilde{\mathbf{v}}\cdot
          (\mathbf{p}^* - \mathbf{p}_\text{base}(\mathbf{q}))}
         {\|\tilde{\mathbf{v}}\|\,
          \|\mathbf{p}^* - \mathbf{p}_\text{base}(\mathbf{q})\|}
  \right) \in [0, 1].
\end{equation}
Like $S_{o,\text{ee}}$, the score depends only on directions, not on absolute distance.
 
The two opposition terms encode the geometry of natural bimanual collaboration to render teleoperation instinctive for the user. 
They are complementary: $S_{o,\text{ee}}$ enforces that the two grippers approach from opposing sides at the workspace level, while $S_{o,\text{base}}$ enforces that the two \emph{platforms} sit in a configuration matching a standard bimanual setup, intuitive from the user's perspective (Figure~\ref{fig:positions}).

 \subsection{Phase 3: grasp candidate generation and ranking}
\label{sec:phase3}
The complement robot then navigates to the Phase~2 base pose and the complement arm captures a close-range RGB--D frame and dense point cloud $\mathcal{P}_3$. 
The VLM is queried a second time on this close-range image to refine the target grasp pixel $\mathbf{u}_3^*$, which is projected to a 3-D point ${}^{\Bframe}\mathbf{p}_3^*$ expressed in the complement base frame. This refined point serves as the \emph{position} of the generated grasp candidates. 
We then sample different grasps and select the optimal grasp based on task need.

\subsubsection{Grasp Orientation Sampling:}
We generate $N$ candidate approach directions $\{\hat{\mathbf{a}}_i\}_{i=1}^{N}$ as a Fibonacci-spiral sampling of the unit
sphere centered at ${}^{\Bframe}\mathbf{p}_3^*$, providing near-uniform coverage of $S^2$ ($N = 64$). For each approach direction we sample $M$ gripper rolls about $\hat{\mathbf{a}}_i$ ($M = 4$), yielding a candidate set $\{\mathbf{T}_C^{(i,j)}\}$ of $N \times M$ gripper poses, each with translation ${}^{\Bframe}\mathbf{p}_3^*$ and an orientation determined by $(\hat{\mathbf{a}}_i, \phi_j)$. Each candidate is checked against the close-range point cloud $\mathcal{P}_3$ using a simplified gripper collision model, and surviving candidates are tested for inverse-kinematic feasibility on the complement arm at the current base pose to return the final set of candidate grasp poses.

\subsubsection{Selection by directional manipulability:}
\label{sec:kinflex}
Each candidate in $\mathcal{T}$ is ranked by how much kinematic flexibility the complement arm retains along the task-relevant directions $\hat{\mathbf{d}}_v, \hat{\mathbf{d}}_\omega$ inferred by the VLM in Phase~1~\cite{yoshikawa1985manipulability}. For a candidate $\mathbf{T}_C^{(i,j)}$, IK gives a configuration $\mathbf{q}^\star$, at which the spatial Jacobian $J(\mathbf{q}^\star)$ in the world-aligned local frame is partitioned into positional and rotational rows, $J = [J_p^\top \; J_r^\top]^\top$. The directional manipulability~\cite{chiu1988taskcompatibility},

\begin{equation}
  w_\text{dir}(\mathbf{q}^\star;\,\hat{\mathbf{d}};\,J_\bullet)
    = \sqrt{\hat{\mathbf{d}}^\top
              J_\bullet J_\bullet^\top \hat{\mathbf{d}}},
  \label{eq:dir_manip}
\end{equation}
is the length of the manipulability ellipsoid in direction $\hat{\mathbf{d}}$, i.e.\ the end-effector velocity available in $\hat{\mathbf{d}}$ per unit joint velocity. The Jacobian partition is task-coupled: $J_\bullet = J_p$ for linear directions, $J_\bullet = J_r$ for angular axes.

When both linear and angular directions are provided, the combined score is
\begin{equation}
  w_\text{cmb} = \beta_v\, w_\text{dir}(\mathbf{q}^\star;\,
                            {}^{\Bframe}\hat{\mathbf{d}}_v;\, J_p)
              + \beta_\omega\, w_\text{dir}(\mathbf{q}^\star;\,
                            {}^{\Bframe}\hat{\mathbf{d}}_\omega;\, J_r),
  \label{eq:combined_manip}
\end{equation}
where the weights $(\beta_v, \beta_\omega)$ are set by the Phase~1 motion type $\mu$: $(1.0, 0.0)$ for \textsc{linear}, $(0.7, 0.3)$ for \textsc{mixed\_linear}, $(0.5, 0.5)$ for \textsc{balanced}, $(0.3, 0.7)$ for \textsc{mixed\_rotational}, and $(0.0, 1.0)$ for \textsc{rotational}.

$w_\text{cmb}$ captures available velocity per unit joint velocity but not where $\mathbf{q}^\star$ sits within the arm's joint limits: a configuration near a limit cannot execute the task, as the binding joint runs out of travel almost immediately. We therefore regularize $w_\text{cmb}$ with a runway penalty $\sigma \in (0, 1]$, obtained by mapping the task direction through the Jacobian at $\mathbf{q}^\star$ to per-joint speeds and taking the minimum time-to-limit $\Delta_\text{limit}$ across all joints,

\begin{equation}
  \sigma = \tanh\!\left(\Delta_\text{limit}\,/\,t \right),
  \quad t = 1.0~\mathrm{s},
  \label{eq:runway_sigma}
\end{equation}

giving the ranking score $w_\text{task} = w_\text{cmb}\cdot\sigma$. The selected grasp is
\begin{equation}
  \mathbf{T}_C^\star
    = \arg\max_{\mathbf{T}_C^{(i,j)} \in \mathcal{T}}
      w_\text{task}(\mathbf{T}_C^{(i,j)}).
\end{equation}


\section{Experimental Setup}
\label{sec:setup}
We design our evaluation to test, whether JOIN (i) yields higher end-to-end success and lower residual operator effort than a task-agnostic geometry-only baseline, and (ii) approaches the reliability of full teleoperation while removing most of the operator's manual involvement. We evaluate the integrated system rather than isolating individual components; per-component ablations and a user study are deferred to future work. 

\subsection{Hardware and Platform}
The anchor is a Kinova Gen3 arm mounted on the right-hand side of a powered wheelchair. The complement is a Hello Robot Stretch~3 mobile manipulator. Phase-1 reasoning uses Gemini Robotics-ER 1.6 \cite{gemini-robotics-er}; kinematics and Jacobians use Pinocchio. During teleoperation, both robots are controlled using two 6-DOF SpaceMouse devices as input. All trials take place on a tabletop in the living room of an apartment environment 

\subsection{Tasks}
We evaluate on four bimanual tasks (Figure~\ref{tab:tasks}). 
In every task, the anchor has committed to a grasp and a prompt describing the task $\tau$ is given. 

\begin{table}[h]
  \centering
  \small
  \begin{tabular}{p{2.5cm}@{\hspace{0.5cm}}p{4.4cm}@{\hspace{0.5cm}}p{4.2cm}@{}}
    \toprule
    Prompt ($\tau$) & Anchor / Complement roles & Success criterion \\
    \midrule
    ``Help lift the box onto the cutting board'' &
      Anchor holds the right side; complement lifts the left side &
      Box lifted and placed onto the cutting board without drop \\
    \vspace{0.1cm}
    ``Help stir the pot with the whisk''  &
      Anchor holds the pot handle; complement grasps the whisk and stirs
      (whisk orientation varies per trial) &
      Whisk inserted and 3 complete stirs without dislodging
      the pot or losing the whisk \\[15pt] 
    ``Help pour the bottle into the mug'' & 
      Anchor holds the receiving mug; complement pours styrofoam balls
      (water surrogate) from the bottle into the mug & 
      All styrofoam balls are poured out of the bottle \\[28pt]
    ``Help open the water bottle'' & 
      Anchor holds the body; complement unscrews the lid &
      Lid fully unscrewed and removed \\
    \bottomrule
  \end{tabular}
  \caption{Task suite. The anchor (wheelchair-mounted Kinova) holds its grasp  fixed before each trial; the complement (Stretch~3) joins from the same starting position each trial.}
  \label{tab:tasks}
\end{table}

\subsection{Protocol}
We run $n=5$ trials per (task, method). In all conditions except full teleoperation, the system runs autonomously up to a proposed complement grasp and then hands control to a single expert teleoperator, who may adjust the grasp before executing the collaborative motion under teleoperation.

\subsection{Baselines}
We compared JOIN against two baselines:

\begin{description}
\item[Teleop (upper bound)] The expert teleoperates the complement arm for the entire task, from its summoned starting pose through task completion, with no autonomous assistance. This is the reliability target we aim to approach while removing operator burden, not a method we aim to beat on success.

\item[AnyGrasp] A task-agnostic baseline representing current state-of-the-art grasping pipeline. The complement selects its joining base pose by proximity to the target object only (no opposition term).
The pipeline utilizes AnyGrasp~\cite{fang2023anygrasp} to generate 6-DOF grasp candidates in the scene. 
Meta's SAM3~\cite{carion2025sam3segmentconcepts} is then used to identify task-relevant object(s) and candidates not grasping those objects are filtered out. 
The highest AnyGrasp-confidence candidate is then selected. 
This baseline represents current practice without task-level reasoning, opposition, or directional manipulability.

\end{description}


\subsection{Metrics}
For each successful trial, we record the following wall-clock quantities, all
measured from the moment the complement is summoned:
\begin{description}
  \item[$t_\text{pre}$, \emph{time to pre-grasp}]: duration the system runs without operator input, ending when control is handed to the operator with a proposed grasp. AnyGrasp and JOIN only; lower is better.
  \item[$t_\text{grasp}$, \emph{grasp-adjustment time}]: duration the operator spends correcting the proposed grasp before beginning the task motion. Defined for AnyGrasp and JOIN only; lower indicates a better proposal.
  \item[$t_\text{post}$, \emph{post-grasp execution time}]: from the end of grasp adjustment to task completion; lower is better.
  \item[$t_\text{total}$, \emph{total task time}]: full duration from summon to completion. 
\end{description}
We additionally report \emph{success rate}, and two task-specific quality metrics: number of \emph{regrasps} in bottle opening, and number of \emph{balls spilled} in pouring.

\section{Results}
\label{sec:results}

\begin{figure}[!t]
  \centering
  \begin{minipage}[t]{0.62\linewidth}
    \strut\vspace*{-\baselineskip}\par
  \centering
  \small
  \begin{tabular}{@{}llccccc@{}}
    \toprule
    & Method & Success & $t_\text{pre}$ & $t_\text{grasp}$ & $t_\text{post}$ & $t_\text{total}$ \\
    \midrule
    \multirow{3}{*}{\rotatebox[origin=c]{90}{Box lift}}
      & Teleop   & 5/5 & --  & --   & 30$\pm$9  & 91$\pm$8  \\
      & AnyGrasp & 4/5 & 78$\pm$19  & 34$\pm$6  & 28$\pm$7 & 139$\pm$26 \\
      & JOIN     & 5/5 & 74$\pm$24  & 15$\pm$4  & 36$\pm$7 & 125$\pm$29 \\
    \cmidrule(lr){2-7}
    \multirow{3}{*}{\rotatebox[origin=c]{90}{\shortstack{Bottle\\open}}}\hspace{0.2cm}
      & Teleop   & 5/5 & --  & --   & 85$\pm$44  & 130$\pm$48 \\
      & AnyGrasp & 5/5 & 67$\pm$8   & 15$\pm$12  & 142$\pm$92 & 224$\pm$86 \\
      & JOIN     & 5/5 & 65$\pm$12  & 10$\pm$4   & 101$\pm$73 & 176$\pm$75 \\
    \cmidrule(lr){2-7}
    \multirow{3}{*}{\rotatebox[origin=c]{90}{Stir pot}}
      & Teleop   & 5/5 & --  & --   & 21$\pm$4  & 75$\pm$10  \\
      & AnyGrasp & 2/5 & 118$\pm$12 & 38$\pm$4  & 26$\pm$13 & 182$\pm$5 \\
      & JOIN     & 4/5 & 78$\pm$25  & 11$\pm$2  & 18$\pm$1  & 107$\pm$26 \\
    \cmidrule(lr){2-7}
    \multirow{3}{*}{\rotatebox[origin=c]{90}{Pour}}
      & Teleop   & 5/5 & --   & --   & 30$\pm$10 & 82$\pm$16  \\
      & AnyGrasp & 3/5 & 104$\pm$50 & 13$\pm$7  & 55$\pm$23 & 172$\pm$62 \\
      & JOIN     & 5/5 & 58$\pm$6   & 11$\pm$5  & 31$\pm$11 & 100$\pm$17 \\
    \midrule
    \multirow{3}{*}{\rotatebox[origin=c]{90}{Overall}}
      & Teleop   & 20/20 & -- & -- & -- & -- \\
      & AnyGrasp & 14/20 &  -- & -- & -- & -- \\
      & JOIN     & 19/20 &  -- & -- & -- & -- \\
    \bottomrule
  \end{tabular}
  \captionof{table}{End-to-end results by task and method. Times are means over completed trials (seconds); $t_\text{pre}$ (autonomous time) and $t_\text{grasp}$ (operator adjustment) are undefined for Teleop, which is fully teleoperated. Complete-after-pre-grasp is $t_\text{total}-t_\text{pre}$; complete-after-grasp is $t_\text{post}$. $\pm$ denotes the standard deviation.}
  \label{tab:results}
  \end{minipage}\hfill
 \begin{minipage}[t]{0.32\linewidth}
   \strut\vspace*{-\baselineskip}\par
    \centering
    \includegraphics[width=\linewidth]{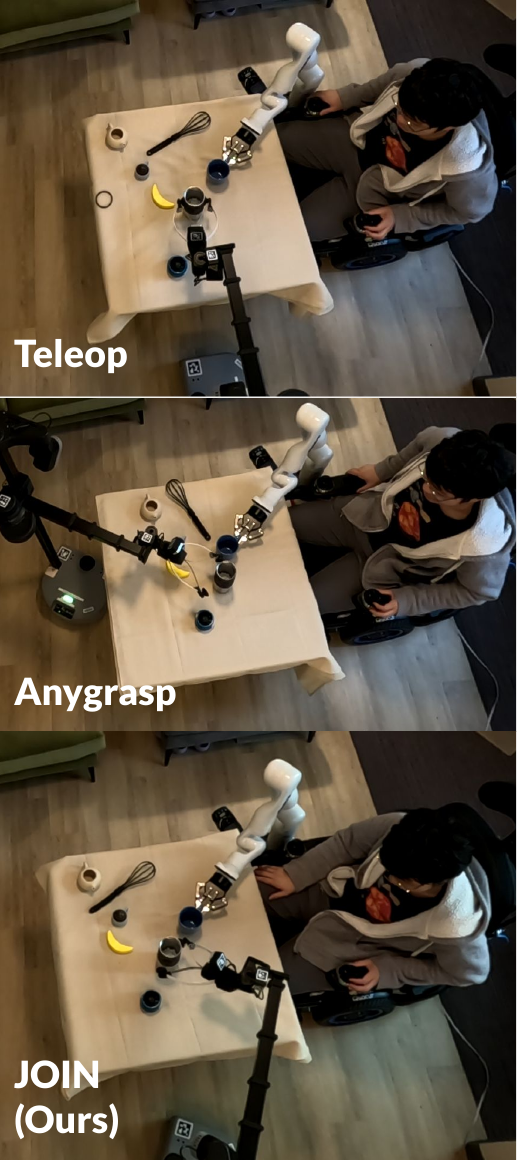}
    \captionof{figure}{Comparison of join and grasp positions for pouring task.}
    \label{fig:positions}
  \end{minipage}
\end{figure}

\subsection{Success and Reliability}
Table~\ref{tab:results} reports success and timing for all four tasks. JOIN succeeds on 19 of 20 trials, almost matching full teleoperation (20/20) and  exceeding the geometry-only AnyGrasp baseline (14/20). All methods perform similarly on the same-object tasks (box lift, bottle open), but diverge on the different-object tasks: the baseline succeeds on only 2/5 (stirring) and 3/5 (pouring), whereas JOIN reaches 4/5 and 5/5. The baseline's failures are informative about what task-agnostic grasp selection misses: on stirring it twice failed to find any valid grasp and once grasped the pot handle rather than the whisk, and on pouring it failed twice when SAM3 could not segment the target object. We emphasize that with five trials per cell these rates are indicative rather than statistically conclusive, and we report them to characterize the system's behavior across the task suite rather than to claim a precise success probability.

\subsection{Operator Effort}
The goal of the bimanual joining problem is to minimize the operator's effort. We evaluate the operator's effort by examining how much time it takes for the operator to complete the teleoperation portion of the tasks.
This excludes the computation and movement time. 
For JOIN and AnyGrasp, this is after the robot joins in the interaction. We examine both $t_\text{grasp}$ (the time the operator spent correcting the end effector before grasping) and $t_\text{post}$ (the time spent completing the task).

JOIN's task-conditioned grasps required less correction than the AnyGrasp baseline for all. On average, the operator spent less time adjusting the grasp -- (15 vs.\ 34\,s on box lift, 10 vs.\ 15\,s on bottle open, 11 vs.\ 38\,s on stirring, and 11 vs.\ 13\,s on pouring). 
Once the operator starts completing the task, the operator in JOIN spends less time in 3 out of the 4 tasks (101 vs.\ 142\,s on bottle opening, 18 vs.\ 26\,s on stirring, and 31 vs.\ 55\,s on pouring). Only in the box-lifting task did the operator take more time to complete the task (36 vs.\ 28s). However, the operator still spent less time in the box lifting task when both $t_\text{grasp}$ and $t_\text{post}$ are combined.
Overall, this shows the proposed grasps by JOIN to be superior to AnyGrasp.


Full teleoperation by an expert remains the fastest method by total task time on every task (91 vs.\ 125\,s on box lift, 130 vs.\ 176\,s on bottle opening, 75 vs.\ 107\,s on stirring, 82 vs.\ 100\,s on pouring). 
This demonstrates that while JOIN outperforms modern autonomous methods, there remains a significant gap between teleoperation and autonomous methods.

\subsection{Task Quality}
For the two tasks with a graded outcome we also examine per-task quality (means over completed trials, lower is better). On bottle opening, JOIN and teleoperation average 2.0 regrasps each, fewer than AnyGrasp's 3.2, consistent with a better-oriented initial grasp on the lid. On pouring, JOIN and teleop spilled little (mean=0.4 for JOIN and mean=0.2 for teleop), where as AnyGrasp split on average 7.3, mostly driven by a single trial that dropped 21 balls from a mis-oriented pour. Both task-related metrics show that JOIN proposed grasps performed similarly to expert-chosen grasps.
 
\section{Discussion}
\label{sec:discussion}
 
We read the reduction in operator effort, rather than raw task time, as JOIN's central practical benefit: a grasp chosen for the task's motion provides a more usable configuration, shifting effort off the operator. Full teleoperation remains the speed ceiling JOIN approaches. JOIN's contribution is the removal of the operator from a significant part of the task: involvement becomes a short grasp correction followed by collaborative execution, with the entire pre-grasp phase handled autonomously. The remaining gap to teleoperation is concentrated in the autonomous overhead and in residual grasp correction, both of which a learned coordination policy is positioned to reduce in future work.

 
 

\subsection{Limitations.}
Our evaluation is deliberately scoped. With five trials per cell the success rates are indicative rather than conclusive, and because we run no ablations our results show that the integrated system works, not that each scoring term is individually necessary. The suite covers four tasks in a static one apartment-style environment, and all trials use a single expert teleoperator who is an author on the paper. We also address only single-step joining.
These choices isolate the system-level question we set out to answer and leave component-level and human-factors questions to future work.

 
 
\section{Conclusion}
\label{sec:conclusion}
 
We introduced \emph{bimanual joining} as a \emph{conditional} form of bimanual manipulation: one arm has committed to a grasp, and the other must complement it while arriving from an independent mobile base. To solve it, we presented \textbf{JOIN}, a three-phase system (plan, drive, grasp) that uses a VLM for task-level reasoning while deferring all 3D quantities to geometric tools. Across four bimanual ADL tasks, JOIN approached the reliability of full teleoperation (19/20 vs.\ 20/20) while outperforming a task-agnostic geometry-only baseline (14/20) and requiring markedly less operator grasp correction. Bimanual joining enables users to perform a class of assistive tasks that single-arm wheelchair systems cannot address; we hope the problem formulation and its decomposition prove useful beyond this specific system.

\section*{Acknowledgments}
This research was funded, in part, by the Advanced Research Projects Agency for Health (ARPA-H) Agreement No. 140D042590012. The views and conclusions contained in this document are those of the authors and should not be interpreted as representing the official policies, either expressed or implied, of the U.S. Government.
\bibliographystyle{splncs03}
\bibliography{references}

@inproceedings{wang2012permma,
  title     = {The Personal Mobility and Manipulation Appliance ({PerMMA}): A robotic wheelchair with advanced mobility and manipulation},
  author    = {Wang, Hongwu and Grindle, Garrett G. and Candiotti, Jorge and Chung, Chengshiu and Shino, Motoki and Houston, Elaine and Cooper, Rory A.},
  booktitle = {Annual International Conference of the IEEE Engineering in Medicine and Biology Society (EMBC)},
  year      = {2012},
}

@inproceedings{gandhi2025bimanualshared,
      author={Gandhi, Rohan and Casado, Fernando Estevez and Demiris, Yiannis},
      booktitle={2025 34th IEEE International Conference on Robot and Human Interactive Communication (RO-MAN)}, 
      title={Toward Shared Control for Mobile Bimanual Manipulation on a Robotic Wheelchair}, 
      year={2025},
      pages={851-856},
      doi={10.1109/RO-MAN63969.2025.11217797}
  }

@article{nanavati2024assistivesurvey,
  title   = {Physically Assistive Robots: A Systematic Review of Mobile and Manipulator Robots That Physically Assist People with Disabilities},
  author  = {Nanavati, Amal and Ranganeni, Vinitha and Cakmak, Maya},
  journal = {Annual Review of Control, Robotics, and Autonomous Systems},
  year    = {2024},
}

@inproceedings{padmanabha2024wearableteleop,
  title     = {Independence in the Home: A Wearable Interface for a Person with Quadriplegia to Teleoperate a Mobile Manipulator},
  author    = {Padmanabha, Akhil and Gupta, Janavi and Chen, Chen and Yang, Jehan and Nguyen, Vy and Weber, Douglas J. and Majidi, Carmel and Erickson, Zackory},
  booktitle = {ACM/IEEE International Conference on Human-Robot Interaction (HRI)},
  year      = {2024},
}

@article{jenamani2024flairfeeding,
  title   = {{FLAIR}: Feeding via Long-horizon {AcquIsition} of Realistic dishes},
  author  = {Jenamani, Rajat Kumar and Sundaresan, Priya and Sakr, Maram and Bhattacharjee, Tapomayukh and Sadigh, Dorsa},
  journal = {arXiv preprint arXiv:2407.07561},
  year    = {2024}
}

@inproceedings{ye2025cartmpc,
  title     = {{CART-MPC}: Coordinating Assistive Devices for Robot-Assisted Transferring with Multi-Agent Model Predictive Control},
  author    = {Ye, Ruolin and Chen, Shuaixing and Yan, Yunting and Yang, Joyce and Ge, Christina and Barreiros, Jose and Tsui, Kate and Silve, Tom and Bhattacharjee, Tapomayukh},
  booktitle = {IEEE International Conference on Robotics and Automation (ICRA)},
  year      = {2025},
}

@inproceedings{jenamani2025enhancing,
    author = {Jenamani, Rajat Kumar and Padmanabha, Akhil and Nanavati, Amal and Cakmak, Maya and Erickson, Zackory and Bhattacharjee, Tapomayukh},
    title = {Enhancing Independence with Physical Caregiving Robots},
    year = {2025},
    publisher = {IEEE Press},
    pages = {1973–1975},
    numpages = {3},
    series = {HRI '25}
}

@inproceedings{grotz2024peract2,
  title     = {{PerAct2}: Benchmarking and Learning for Robotic Bimanual Manipulation Tasks},
  author    = {Grotz, Markus and Shridhar, Mohit and Asfour, Tamim and Fox, Dieter},
  booktitle = {arXiv preprint arXiv:2407.00278},
  year      = {2024}
}

@article{jiang2025decoupledbimanual,
  title   = {Rethinking Bimanual Robotic Manipulation: Learning with Decoupled Interaction Framework},
  author  = {Jiang, Jian-Jian and Wu, Xiao-Ming and He, Yi-Xiang and Zeng, Ling-An and Wei, Yi-Lin and Zhang, Dandan and Zheng, Wei-Shi},
  journal = {arXiv preprint arXiv:2511.02215},
  year    = {2025}
}

@article{im2025twinvla,
  title   = {{TwinVLA}: Data-Efficient Bimanual Manipulation with Twin Single-Arm Vision-Language-Action Models},
  author  = {Im, Hokyun and Jeong, Euijin and Fu, Jianlong and Kolobov, Andrey and Lee, Youngwoon},
  journal = {arXiv preprint arXiv:2511.04860},
  year    = {2025}
}

@article{hahne2026taskaware,
  title   = {Task-Aware Bimanual Affordance Prediction via {VLM}-Guided Semantic-Geometric Reasoning},
  author  = {Hahne, Fabian and Prasad, Vignesh and Chalvatzaki, Georgia and Peters, Jan and Kshirsagar, Alap},
  journal = {arXiv preprint arXiv:2604.08726},
  year    = {2026}
}

@article{chen2026vlmsfd,
  title   = {{VLM-SFD}: {VLM}-Assisted Siamese Flow Diffusion Framework for Dual-Arm Cooperative Manipulation},
  author  = {Chen, Jiaming and Jiang, Yiyu and Huang, Aoshen and Li, Yang and Pan, Wei},
  journal = {IEEE Robotics and Automation Letters},
  year    = {2026}
}

@article{heidinger2025handedafforder,
  title   = {2{HandedAfforder}: Learning Precise Actionable Bimanual Affordances from Human Videos},
  author  = {Heidinger, Marvin and Jauhri, Snehal and Prasad, Vignesh and Chalvatzaki, Georgia},
  journal = {arXiv preprint arXiv:2507.00500},
  year    = {2025}
}

@inproceedings{liu2024okrobot,
  title     = {{OK-Robot}: What Really Matters in Integrating Open-Knowledge Models for Robotics},
  author    = {Liu, Peiqi and Orru, Yaswanth and Vakil, Jay and Paxton, Chris and Shafiullah, Nur Muhammad Mahi and Pinto, Lerrel},
  booktitle = {Robotics: Science and Systems (RSS)},
  year      = {2024}
}

@inproceedings{liu2025dynamem,
  title     = {{DynaMem}: Online Dynamic Spatio-Semantic Memory for Open World Mobile Manipulation},
  author    = {Liu, Peiqi and Guo, Zhanqiu and Warke, Mohit and Chintala, Soumith and Paxton, Chris and Shafiullah, Nur Muhammad Mahi and Pinto, Lerrel},
  booktitle = {IEEE International Conference on Robotics and Automation (ICRA)},
  year      = {2025},
}

@article{shi2025hirobot,
  title   = {Hi Robot: Open-Ended Instruction Following with Hierarchical Vision-Language-Action Models},
  author  = {Shi, Lucy Xiaoyang and Ichter, Brian and Equi, Michael and Ke, Liyiming and Pertsch, Karl and Vuong, Quan and Tanner, James and Walling, Anna and Wang, Haohuan and Fusai, Niccolo and Li-Bell, Adrian and Driess, Danny and Groom, Lachy and Levine, Sergey and Finn, Chelsea},
  journal = {arXiv preprint arXiv:2502.19417},
  year    = {2025}
}

@article{chen2025emos,
  title   = {{EMOS}: Embodiment-aware Heterogeneous Multi-robot Operating System with {LLM} Agents},
  author  = {Chen, Junting and Yu, Checheng and Zhou, Xunzhe and Xu, Tianqi and Mu, Yao and Hu, Mengkang and Shao, Wenqi and Wang, Yikai and Li, Guohao and Shao, Lin},
  journal = {arXiv preprint arXiv:2410.22662},
  year    = {2025}
}

@inproceedings{mandi2024roco,
  title     = {{RoCo}: Dialectic Multi-Robot Collaboration with Large Language Models},
  author    = {Mandi, Zhao and Jain, Shreeya and Song, Shuran},
  booktitle = {IEEE International Conference on Robotics and Automation (ICRA)},
  year      = {2024},
}

@article{yoshikawa1985manipulability,
  title   = {Manipulability of Robotic Mechanisms},
  author  = {Yoshikawa, Tsuneo},
  journal = {The International Journal of Robotics Research},
  volume  = {4},
  number  = {2},
  pages   = {3--9},
  year    = {1985},
  doi     = {10.1177/027836498500400201}
}

@article{chiu1988taskcompatibility,
  title   = {Task Compatibility of Manipulator Postures},
  author  = {Chiu, Stephen L.},
  journal = {The International Journal of Robotics Research},
  volume  = {7},
  number  = {5},
  pages   = {13--21},
  year    = {1988},
  doi     = {10.1177/027836498800700502}
}

@inproceedings{zacharias2007capability,
  title     = {Capturing Robot Workspace Structure: Representing Robot Capabilities},
  author    = {Zacharias, Franziska and Borst, Christoph and Hirzinger, Gerd},
  booktitle = {IEEE/RSJ International Conference on Intelligent Robots and Systems (IROS)},
  pages     = {3229--3236},
  year      = {2007},
  doi       = {10.1109/IROS.2007.4399105}
}

@inproceedings{vahrenkamp2013reachabilityinversion,
  title     = {Robot Placement based on Reachability Inversion},
  author    = {Vahrenkamp, Nikolaus and Asfour, Tamim and Dillmann, R\"udiger},
  booktitle = {IEEE International Conference on Robotics and Automation (ICRA)},
  pages     = {1970--1975},
  year      = {2013},
  doi       = {10.1109/ICRA.2013.6630839}
}

@article{jauhri2022reachabilitypriors,
  title   = {Robot Learning of Mobile Manipulation with Reachability Behavior Priors},
  author  = {Jauhri, Snehal and Peters, Jan and Chalvatzaki, Georgia},
  journal = {IEEE Robotics and Automation Letters},
  volume  = {7},
  number  = {3},
  pages   = {8399--8406},
  year    = {2022},
  doi     = {10.1109/LRA.2022.3188109}
}

@article{shen2026bipremanip,
  title   = {{BiPreManip}: Learning Affordance-Based Bimanual Preparatory Manipulation through Anticipatory Collaboration},
  author  = {Shen, Yan and Jiang, Feng and He, Zichen and Li, Xiaoqi and Liu, Yuchen and Li, Zhiyu and Wu, Ruihai and Dong, Hao},
  journal = {arXiv preprint arXiv:2603.21679},
  year    = {2026}
}

@article {fang2023anygrasp,
  title   = {{AnyGrasp}: Robust and Efficient Grasp Perception in Spatial and Temporal Domains},
  author  = {Fang, Hao-Shu and Wang, Chenxi and Fang, Hongjie and Gou, Minghao and Liu, Jirong and Yan, Hengxu and Liu, Wenhai and Xie, Yichen and Lu, Cewu},
  journal = {IEEE Transactions on Robotics},
  volume  = {39},
  number  = {5},
  pages   = {3929--3945},
  year    = {2023},
  doi     = {10.1109/TRO.2023.3281153}
}

@misc{gemini-robotics-er,
  author       = {{Google DeepMind}},
  title        = {Gemini Robotics-ER 1.6},
  howpublished = {Model Card},
  year         = {2026},
  url          = {https://deepmind.google},
}

@INPROCEEDINGS{padir2015personalized,
  author={Padır, Taşkın},
  booktitle={2015 37th Annual International Conference of the IEEE Engineering in Medicine and Biology Society (EMBC)}, 
  title={Towards personalized smart wheelchairs: Lessons learned from discovery interviews}, 
  year={2015},
  volume={},
  number={},
  pages={5016-5019},
  doi={10.1109/EMBC.2015.7319518}
  }

@inproceedings{kelecstemur2019system,
 title={System architecture for autonomous mobile manipulation of everyday objects in domestic environments},
 author={Kele{\c{s}}temur, Tarik and Yokoyama, Naoki and Truong, Joanne and Allaban, Anas Abou and Padir, Ta{\c{s}}kin},
 booktitle={Proceedings of the 12th ACM International Conference on PErvasive Technologies Related to Assistive Environments},
 pages={264--269},
 year={2019}
}

@misc{carion2025sam3segmentconcepts,
      title={SAM 3: Segment Anything with Concepts},
      author={Nicolas Carion and Laura Gustafson and Yuan-Ting Hu and Shoubhik Debnath and Ronghang Hu and Didac Suris and Chaitanya Ryali and Kalyan Vasudev Alwala and Haitham Khedr and Andrew Huang and Jie Lei and Tengyu Ma and Baishan Guo and Arpit Kalla and Markus Marks and Joseph Greer and Meng Wang and Peize Sun and Roman Rädle and Triantafyllos Afouras and Effrosyni Mavroudi and Katherine Xu and Tsung-Han Wu and Yu Zhou and Liliane Momeni and Rishi Hazra and Shuangrui Ding and Sagar Vaze and Francois Porcher and Feng Li and Siyuan Li and Aishwarya Kamath and Ho Kei Cheng and Piotr Dollár and Nikhila Ravi and Kate Saenko and Pengchuan Zhang and Christoph Feichtenhofer},
      year={2025},
      eprint={2511.16719},
      archivePrefix={arXiv},
      primaryClass={cs.CV},
      url={https://arxiv.org/abs/2511.16719},
}

\end{document}